\documentclass[twocolumn,english,10pt,a4paper]{article}

\usepackage[T1]{fontenc} 
\usepackage{eurosis}
\usepackage{url}
\usepackage{graphicx}

\title{\textbf{EMOTION: APPRAISAL-COPING MODEL FOR THE ``CASCADES'' PROBLEM}}
\author{Karim Mahboub$^{(1)}$, Evelyne Cl\'ement$^{(2)}$, Cyrille Bertelle$^{(1)}$ \& V\'eronique Jay$^{(1)}$ \\ ~ \\
	\small{$^{(1)}$LITIS Laboratory, University of Le Havre} \\ 
	\small{25, rue Philippe Lebon, BP 540 76058 Le Havre Cedex, France} \\
	\small{\texttt{Karim.Mahboub@litislab.eu}} \\ 
	\small{\texttt{Cyrille.Bertelle@litislab.eu}} \\ 
	\small{\texttt{Veronique.Jay@litislab.eu}} \\ ~ \\
	\small{$^{(2)}$Psy.CO Laboratory, University of Rouen} \\ 
	\small{Rue Lavoisier 76821 Mont-Saint-Aignan, France} \\
	\small{\texttt{Evelyne.Clement@univ-rouen.fr}}
	}
\date{ }

\begin{document}

\maketitle

\begin{abstract}
Modelling emotion has become a challenge nowadays. Therefore, several models 
have been produced in order to express human emotional activity. However, only a 
few of them are currently able to express the close relationship existing 
between emotion and cognition. An appraisal-coping model is presented here, with 
the aim to simulate the emotional impact caused by the evaluation of a 
particular situation (appraisal), along with the consequent cognitive reaction 
intended to face the situation (coping). This model is applied to the 
``Cascades'' problem, a small arithmetical exercice designed for ten-year-old 
pupils. The goal is to create a model corresponding to a child's behaviour when
solving the problem using his own strategies.

\end{abstract}

%
%
\section*{INTRODUCTION}

As the study of emotion is becoming crucial today, in several fields of study 
such as neurology or psychology, computer science is getting more and more 
involved in the process of finding new models for representing emotions. Since 
the middle of the 19th century, psychologists, biologists, but also neurologists 
have tried to produce models designed to unravel the emotional processes. 
Scientists like Bechara and Damasio \cite{Bechara2000} have even proved that 
human emotional activity has an indispensable impact on decision making.

The aim of this article is to globally draw the picture of the latest models of
emotion in computer science, starting from a psychologist point of view. The 
appraisal-coping model will be presented more precisely, as well as an existing
application example. Afterwards, we will present a small toy problem 
illustrating a decision making problem, using the appraisal-coping strategy. 
Eventually, a small conclusion will introduce future aspects to be developped 
and other prospects.

%
%
\section*{STATE OF THE ART}

From the most recent approaches of emotion modelling, two major cognitive types
of research models have been developed: hierarchical and componential models
(see Baudic \& Duchamp, 2006 \cite{Baudic2006}). 

According to the hierarchical approach, emotions have their origin from early 
stages of development. Indeed, emotions are hierarchically organised with 
numerous discrete emotions at a basic level and emotional dimensions at a higher 
level. Fundamental emotions like thirst or fear are elements intended to build 
more sophisticated emotions, ending up with very complex feelings such as 
jealousy or pride \cite{Denton2006}. Emotions at the basic level have an important 
adaptive function and are directly linked to the body stimuli and effectors.

In componential models, emotions have qualitatively different facets 
\cite{Scherer2001a}. The so-called "emotional response triad" is composed of the 
three main components for the emotion production: subjective experience, 
peripheral physiological responses and motor expression, to which some theorists 
include two other components, cognitive and motivational. The componential 
approach deals with the relative role assigned to each of these components. 
Then, emotions are created by stepping through all parts of the process, from the 
cognitive perception, until the actual response. Lazarus \cite{Lazarus1982} and 
Scherer \cite{Scherer1984} are usually associated with this approach.

%
%
\section*{APPRAISAL THEORIES}

Appraisal theories suggest that emotion is the result of underlying mechanisms 
including the subjective evaluation of the significance of a situation and its 
organism circumstances (appraisal), and the coping mechanisms that guide and 
provide adaptive responses (Frijda, 1986 \cite{Frijda1986}; Lazarus, 1991 
\cite{Lazarus1991}; Scherer, 1984 \cite{Scherer1984}; Scherer, Schorr \& 
Johnstone, 2001 \cite{Scherer2001c}; Smith \& Lazarus, 1990 \cite{Smith1990}, 
1993 \cite{Smith1993}). As noted by Gratch and Marsella \cite{Gratch2004}: 
"Appraisal theories posit that events do not have significance in of themselves, 
but only by virtue of their interpretation in the context of an individual's 
beliefs, desires, intentions and abilities" (Gratch \& Marsella, 2004, p. 273). 
The significance of an event is supposed to be evaluated on a number of criteria 
such as its relevance for one's well-being, its conduciveness for one's plans 
and goals, and the ability to cope with such consequences.

In the framework of the Scherer's Component Process Model (Scherer 1984 
\cite{Scherer1984}, 2001 \cite{Scherer2001c}), Sander, Grandjean \& Scherer (2005 
\cite{Sander2005}) describe emotion ``as an episode of interrelated, 
synchronized changes in the states of all or most of the five organismic 
subsystems\footnote{Organismic subsystem (and their major substata are the 
following): Information processing (Central Nervous System -- CNS), Support (CNS, 
Neuro-Endocrine System, Autonomic Nervous System), Executive (CNS), Action 
(Somatic Nervous System), Monitor (CNS), from Sander, Grandjean \& Scherer (2005,
\cite{Sander2005}).} 
in response to the evaluation of an external or internal stimulus event as 
relevant to major concerns of the organism.", (p.318). From this point of view, 
rather than static and basic states of the organism (e.g. Ekman, 1984 
\cite{Ekman1984}; Izard, 1971 \cite{Izard1977}), emotions are a dynamic process 
whose components are the cognitive component which function is the evaluation of 
objects and events, the peripheral efference component which regulates the 
system, the motivational component which prepares and guides the actions, the 
motor expression component which steadies communication of reaction and 
behavioural intention, and the subjective feeling component which monitors the 
internal state and environment interaction. In other respects, this model 
postulates that changes in one subsystem will tend to elicit related changes in 
other subsystems. 

%
%
\section*{APPRAISAL-COPING EXAMPLE}

\begin{figure*}[ht]
	\begin{center}
		\includegraphics[scale=0.50]{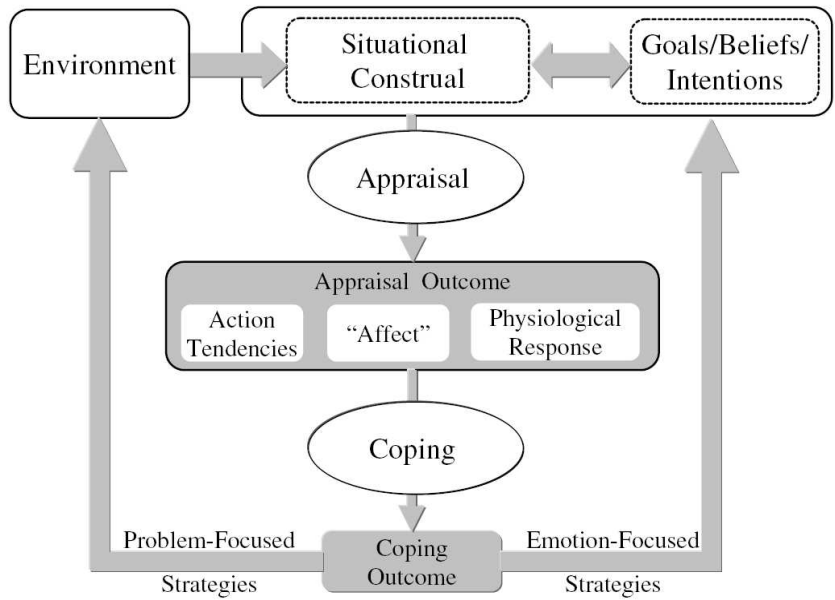}
		\caption{The cognitive-motivational-emotive system. Adapted from 
		Smith an Lazarus (1990, \cite{Smith1990}).\label{fig:modelAC}}
	\end{center}
\end{figure*}

Based on the appraisal-coping approach, several new models have been conceived
(see figure \ref{fig:modelAC}). Gratch \& Marsella \cite{Gratch2004} have 
produced a domain independent model using cognitive maps. This model is intended 
to manipulate appraisal variables to analyse the present and past situations, 
and to design the future decisions to make and the coping strategies to adopt. 
Generally speaking, the appraisal-coping approach offers a very precise model of 
cognitive and emotional processes in decision making (previous works have been
carried out on this subject, see \cite{Mahboub2006b}).

On the one hand, the environment is appraised with respect to one's goals and 
beliefs. This evaluation is realised through a certain number of variables 
defining the different appraising dimensions to be taken into account, such as 
the relevance (\textit{Does the event require attention or adaptive reaction?}) 
or unexpectedness (\textit{Was the event predicted from past knowledge?}). 

On the other hand, one has to cope with the situation appraised before by using
coping strategies. These different strategies offer a great range of 
possibilities, from the perfect control of the situation until the total 
resignation, when facing the problem. The coping outcome alters the 
person-environment relationship not only by modifying the environment itself 
related to the problem, but also by changing the interpretation and willing 
through emotional aspects.

\begin{figure*}[ht]
	\begin{center}
		\includegraphics[scale=0.50]{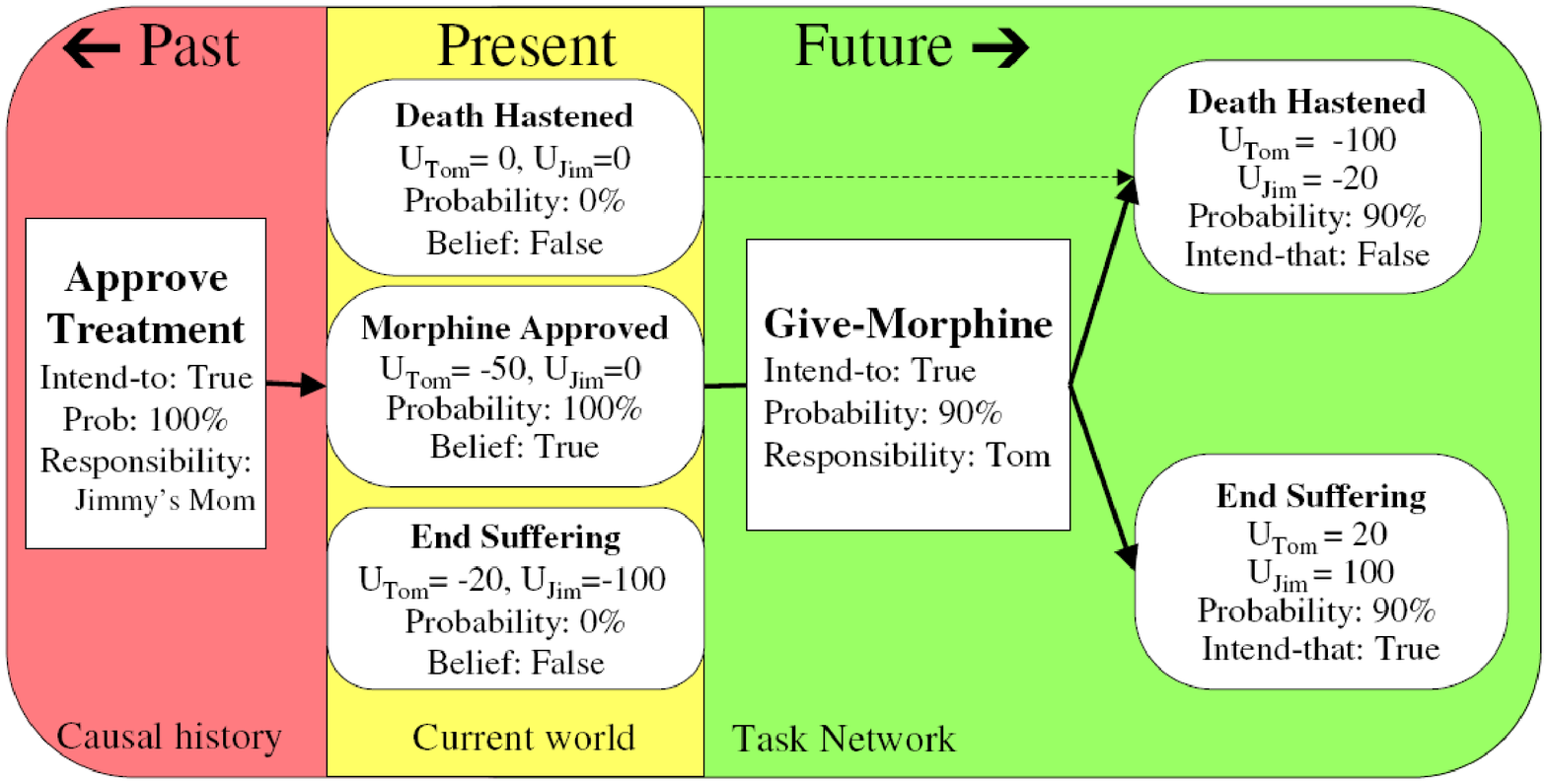}
		\caption{Dr. Tom's causal interpretation at the end of the scenario.
		\label{fig:causalMap}}
	\end{center}
\end{figure*}

In order to represent the information about the situation, Gratch \& Marsella
\cite{Gratch2004} use causal maps. In the following scenario (figure 
\ref{fig:causalMap}), an oncologist, Dr. Tom, is supposed to help an 
eleven-year-old boy, Jimmy, for his stage 4 inoperable cancer, either by giving 
him morphine (which relieves the pain but hastens death) or leaving him suffer 
(and letting him prolong his life). A causal map (see figure 
\ref{fig:causalMap}) represents the past and present situations, as well as the 
possible decisions to make and their expected consequences.

This causal map can evolve with the modification of the parameters. For example,
$U_{Jim}$ is how Dr. Tom appreciates Jim's desirability for the corresponding 
event. This value can be updated after the coping process, dealing with 
emotional aspects. Once Dr. Tom re-appraises the consequences of his decision
(by decreasing the probability parameter of the "Death hastened" event), 
he accepts to give Jim morphine, following Jim's mother request.

However, the major problem when using this technique is the storage of the 
information taken from the environment. In fact, we have two options.

The first option is to put all the information we need explicitly on the object 
to be evaluated. For instance, if an individual is watching a photograph of his 
wife and kids, the variables indicating pleasantness or pride are to be 
described on the photograph. With this method, each object in the environment is 
clearly identified as pleasant or mysterious or annoying, etc. The 
implementation is therefore easier, and the possibility of interacting with the 
emotional representation of the object is hence trivial. The main problem is a 
lack of flexibility, especially if more than one individual has to evaluate the 
same object.

The second option is the internal storage of the objects data, directly into the 
brain, using a memory strategy (Tulving et al., 1972 \cite{Tulving1972}). With 
this technique, memory is divided into several categories, each of which stores 
different kind of information. On the one hand, the long-term memory system is 
composed of the episodic memory (i.e. it refers to knowledge of episodes and 
facts that can be consciously recalled and related) and the semantic memory 
(underlying absolute knowledge and language; semantic memory is 
context-independent). On the other hand, the short-term memory, also known as 
working memory stores the current context-related data.

For each of the above options, emotion is supposed to be triggered from the 
appraisal processes. Indeed, the evaluation activity requires knowledge, and 
emotion is part of the memory processes of encoding, storage and retrieval 
(Tulving \& Thomson, 1973 \cite{Tulving1973}). For modelling purpose, we assume
that memory is split into three categories :

\begin{itemize}
	\item The semantic or factual memory which stores the global knowledge of
	the world and the information considered to be facts, like \textit{"Paris is
	the capital of France"}.
	\item The episodic or autobiographical memory which contains the personal 
	events that happened in an individual's life. This type of memory is 
	strongly linked with a spatio-temporal context.
	\item The working memory is the current dynamic representation that an
	individual has in mind when solving a particular problem.
\end{itemize}

Each time a memory element is encoded (i.e. added into the brain), it is stored
along with the current emotional learning context $e_{t_0}$. Later on, when the 
information is required and must be found in the memory, the current emotional 
context $e_t$ will be compared to the former emotional learning context related
to the required memory item. If the old emotional context equals the current one
(i.e. if $e_{t_0} = e_t$) the retrieval mechanism will be facilitated. On the 
contrary, if the two values are different ($e_{t_0} \ne e_t$), the retrieval
process will be made more difficult.

%
%
\section*{THE ``CASCADES" PROBLEM}

The ``Cascades" problem is a puzzle-like situation in which the goal is to fill 
up the grid with numbers according to the following instruction: ``Each box 
contains the sum of the numbers situated above it. Look for the missing numbers 
in the grid".

\begin{figure*}[ht]
	\begin{center}
		\includegraphics[scale=0.25]{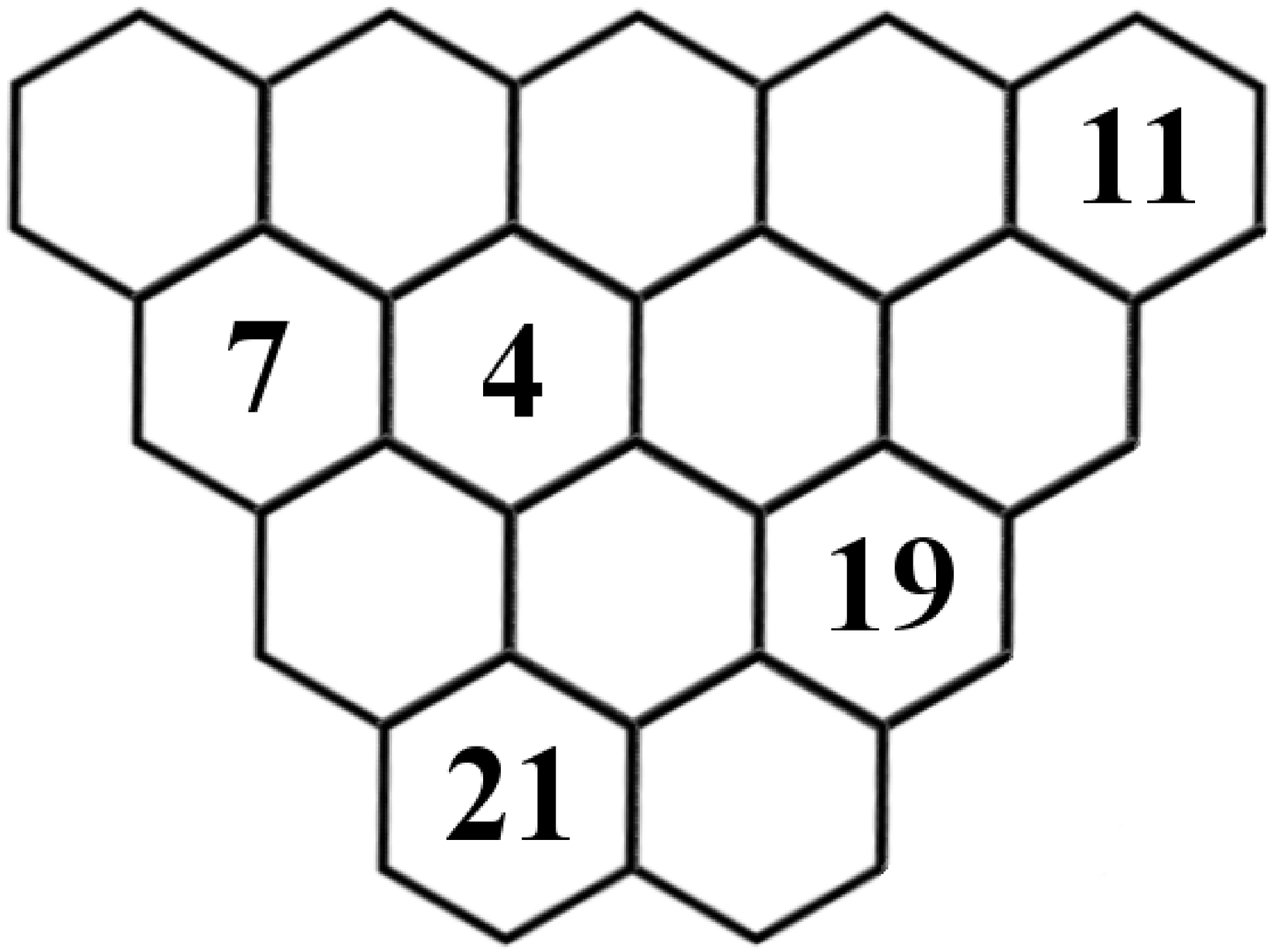} ~~~~~~~~
		\includegraphics[scale=0.25]{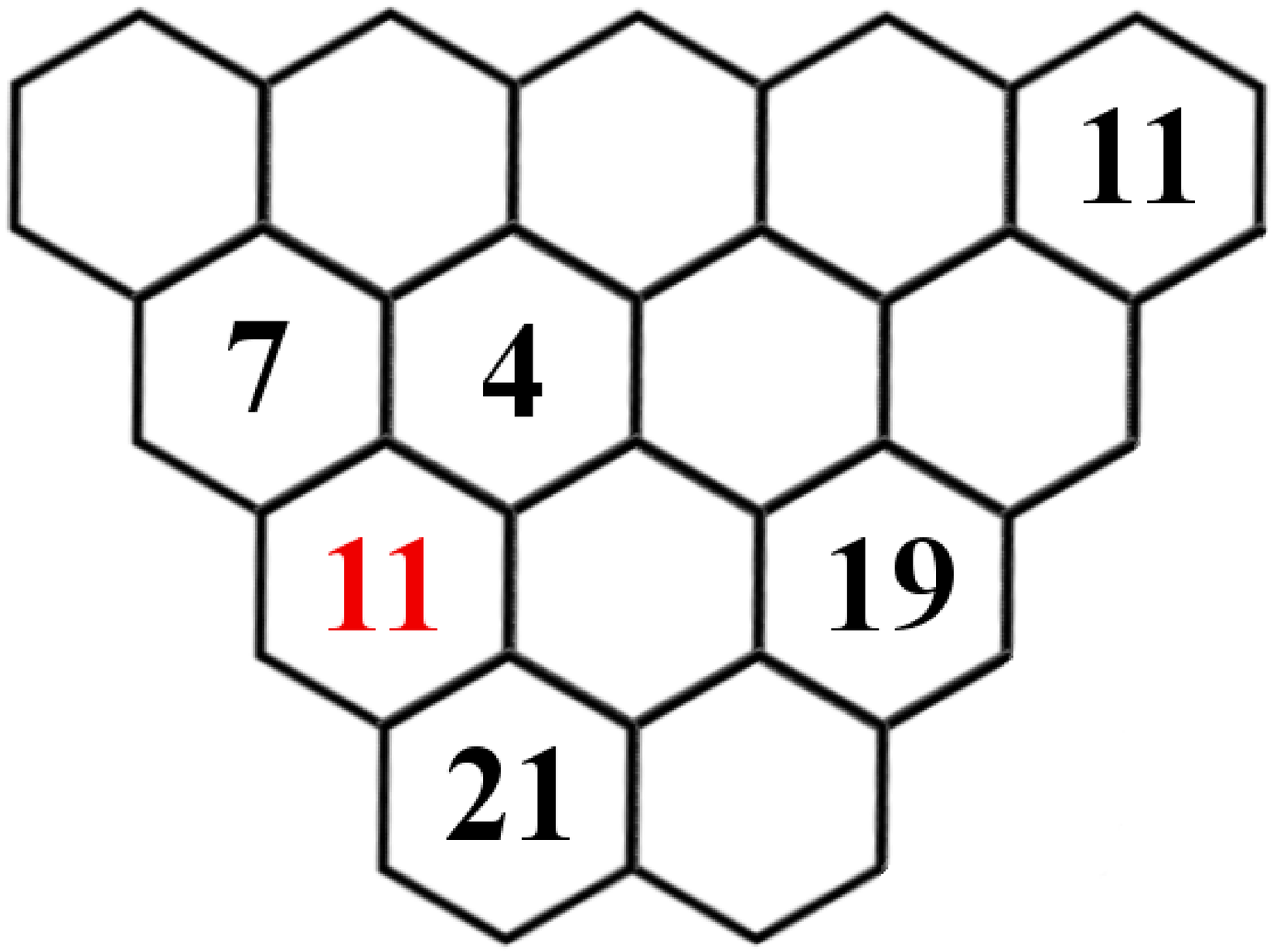}
		\caption{The ``\textit{Cascades}'' initial grid and the first solving step.\label{fig:Cascades}}
	\end{center}
\end{figure*}

The initial state and the first solving step are presented hereafter (figure 
\ref{fig:Cascades}).

\begin{figure*}[ht]
	\begin{center}
		\includegraphics[scale=0.20]{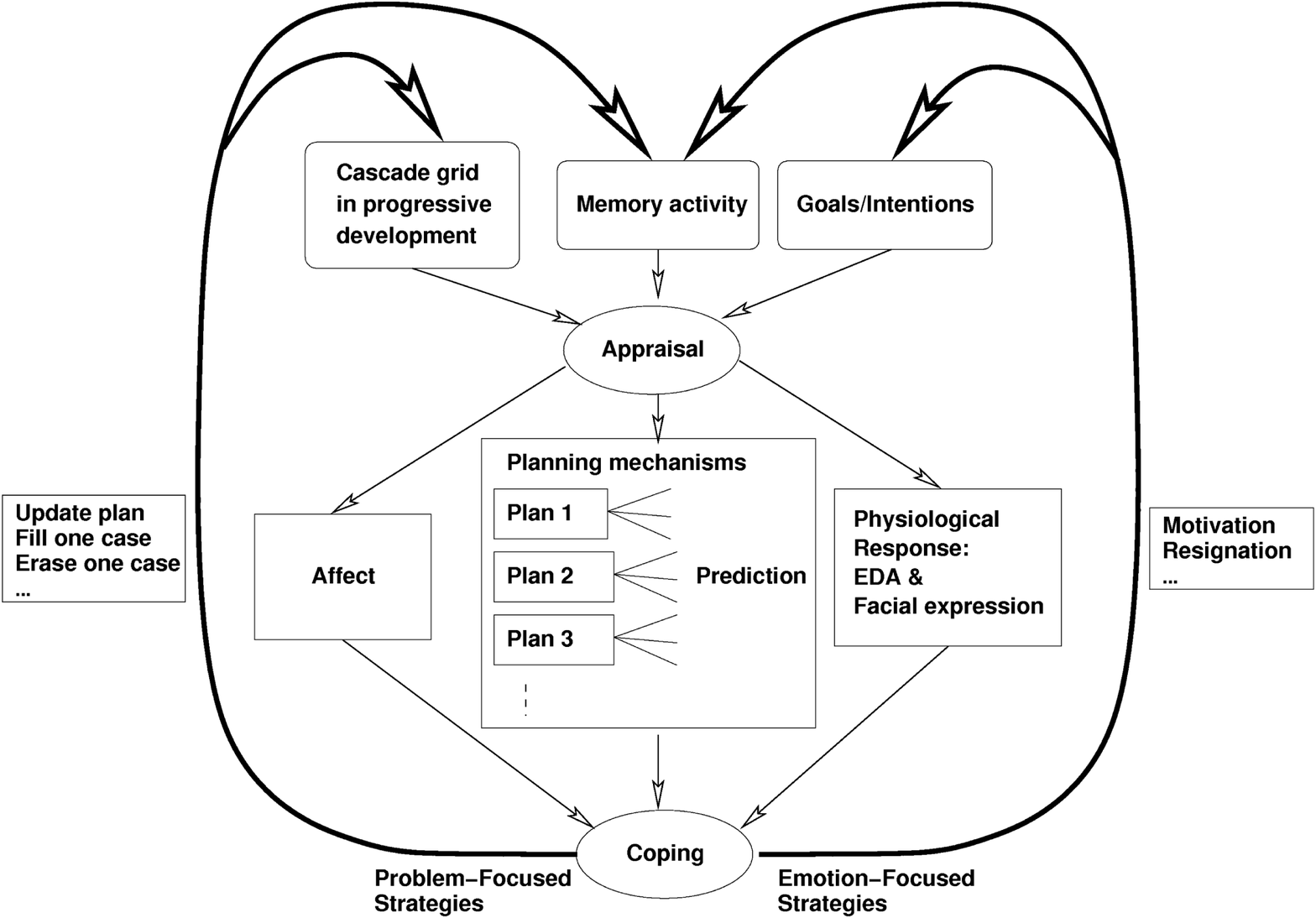}
		\caption{The appraisal-coping model applied to the "Cascades" problem.
		\label{fig:ac-cascade}}
	\end{center}
\end{figure*}

Ten-year-old children will be invited to solve the problem. In order to study 
how the emotions are elicited over the course of the problem, emotional 
manifestations will be recorded without interruption during the problem-solving 
activity. In line with the work of Cl\'ement \& Duvallet (2007 
\cite{Clement2007b}), we will focus on two kinds of response components: the 
physiological -- electrodermal activity -- and the expressive -- facial 
expressions -- components.

Although the definition of emotions remains controversial, some researchers 
distinguish emotions and other related notions as mood or personality traits on 
the basis of their behavioural time course and intensity: emotions are defined 
as short-lived behavioural dispositions, moods are of longer duration and lower 
intensity, while personality traits reflect relatively stable behavioural 
tendencies.

Nevertheless, it is generally assumed that emotions may be evaluated by three 
kinds of responses: the physiological responses which the electrodermal activity 
and the heart rate are the most widely used, the expressive responses including 
facial, vocal, gestural and postural expressions and the subjective responses 
based in part on verbal report (see Bauer, 1998 \cite{Bauer1998}; Boehner, 
DePaula, Dourish, \& Sengers, 2007 \cite{Boehner2007}, for critical reviews). 

Concerning the physiological component, and in particular the electrodermal 
activity, the study of Pecchinenda and Smith (1996) provides psychological 
significance of spontaneous skin conductance activity. In that work, 
participants were given to solve a set of anagrams which difficulty was 
manipulated by both the objective difficulty (easy, moderately, difficult, 
extremely difficult) and the amount of time available to solve the problems (30 
vs. 120 seconds). The authors demonstrate that the skin conductance activity 
during problem solving is correlated to the appraisals of coping potential: in a 
difficult problem, appraisals of coping potential based on self-report are 
especially low and produce selective disengagement of the task, yielding reduced 
skin conductance activity. The spontaneous electrodermal activity is interpreted 
as reflecting task engagement (Pecchinenda, 2001 \cite{Pecchinenda2001}). 

Moreover, the results of Cl\'ement and Duvallet (2007 \cite{Clement2007b}) 
support the idea that skin conductance activity is a convergent measure of 
appraisal-related processes and that facial expressions reflect the appraisals 
of the events according to their conduciveness for the goal (Kaiser \& Wehrle, 
2001 \cite{Kaiser2001}; Scherer, 1999 \cite{Scherer1999}; Smith, 1991 
\cite{Smith1991}; Smith \& Scott, 1997 \cite{Smith1997}).

%
%
\section*{THE APPRAISAL-COPING MODEL}

The appraisal-coping model adaptation to the "Cascades" problem (see figure 
\ref{fig:ac-cascade}) allows us to analyse in a more accurate way the children's 
cognitive emotional activities. The appraisal step deals with the evaluation and 
the prediction of the plan selected by the child in order to solve the problem. 
According to this appraisal step, the child will fill in a hexagon, following 
his selected plan, or correct a previous result, or if the appraisal step leads 
to a bad evaluation, he will change his plan (we observe that the exercice 
instructions are interpreted in many different ways, especially for children 
with school difficulties).

The coping strategy is the actual decision a child will take with respect to his
previous choices. The coping process is usually accompanied by an emotional 
reaction which depends on the appraisal consequences. For instance, if the child
decides to reinforce his strategy, his emotional state will be positive. On the 
contrary, if he continually changes his plans, the situation will end up with a
progressive disengagement for the task.


%
%
\section*{CONCLUSION AND PROSPECTS}

Contrary to the classical models which study cognition in a "cold" way, 
independently from any emotional process, we are now trying to unravel the 
mysteries underlying the emotion-cognition interaction. This new challenge makes
researchers produce a new generation of cognitive models, based on more accurate
systems, such as the appraisal-coping approach.

The multidisciplinary work carried out so far allows us a better understanding 
of emotion mechanisms, by bringing out two complementary approaches: 

\begin{itemize}
	\item The produced model aims to analyse the experimental data.
	\item The experimental data come to strengthen or question the existing 
	model.
\end{itemize}

As a future work, the strategies selected in the coping process are to be 
examined more precisely, in order to extract cognitive-emotional individual 
profiles, and especially when it comes to help pupils with school difficulties.


\end{document}